# Colloquial Persian POS (CPPOS) Corpus: A Novel Corpus for Colloquial Persian Part of Speech Tagging

Leyla Rabiei, Iran Telecommunication Research Center (ITRC), Tehran, Iran, l.rabiei@itrc.ac.ir
Farzaneh Rahmani, Iran Telecommunication Research Center (ITRC), Tehran, Iran, rahmani@itrc.ac.ir
Mohammad Khansari, m.khansari@ut.ac.ir, Faculty of New Sciences and Technologies, University of Tehran, Tehran, Iran
Zeinab Rajabi[*], Iran Telecommunication Research Center (ITRC), Tehran, Iran, z.rajabi@itrc.ac.ir
Moein Salimi, Sharif University of Technology, moein.salimi73@sharif.edu

**Abstract**

**Introduction**: Part-of-Speech (POS) Tagging, the process of classifying words into their respective parts of speech (e.g., verb or noun), is essential in various natural language processing applications. POS tagging is a crucial preprocessing task for applications like machine translation, question answering, sentiment analysis, etc. However, existing corpora for POS tagging in Persian mainly consist of formal texts, such as daily news and newspapers. As a result, smart POS tools, machine learning models, and deep learning models trained on these corpora may not perform optimally for processing colloquial text in social network analysis.

**Method:** This paper introduces a novel corpus, "Colloquial Persian POS" (CPPOS), specifically designed to support colloquial Persian text. The corpus includes formal and informal text collected from various domains such as political, social, and commercial on Telegram, Twitter, and Instagram more than 520K labeled tokens. After collecting posts from these social platforms for one year, special preprocessing steps were conducted, including normalization, sentence tokenizing, and word tokenizing for social text. The tokens and sentences were then manually annotated and verified by a team of linguistic experts. This study also defines a POS tagging guideline for annotating the data and conducting the annotation process.

**Results:** To evaluate the quality of CPPOS, various deep learning models, such as the RNN family, were trained using the constructed corpus. A comparison with another well-known Persian POS corpus named "Bijankhan" and the Persian Hazm POS tool trained on Bijankhan revealed that our model trained on CPPOS outperforms them. With the new corpus and the BiLSTM deep neural model, we achieved a 14% improvement over the previous dataset.

**Keywords**: Part of Speech Tagging, Natural Language Processing, Colloquial Text Processing, Colloquial POS Corpus.

## 1. Introduction

Part-of-Speech (POS) tagging involves assigning lexical tags to words and symbols in the text [1].

These tags indicate the syntactic roles of symbols and words based on their context and sentence structure. POS tagging is fundamental in numerous natural language processing (NLP) applications, including question-answering, machine translation, information retrieval, and text summarization [2]. POS tags provide insights into the structural characteristics of lexical terms within a sentence or text, enabling assumptions about semantics. They also find applications in Named Entity Recognition, Co-reference Resolution, and Speech Recognition.

---

[*]Corresponding author

For instance, Chotriat et al. utilized POS tags for question classification in Thai [3] to enhance classification accuracy. In [4], a recommender system was proposed for social networks to categorize content and provide suggestions based on overlapping interests. They employed POS tags for each word to assign values and obtain information representations to identify shared interests. In Persian Sentiment Analysis, most studies use POS tags to classify sentiments [5, 6]. POS tagging helps to disambiguate words; words with an adjective role in a sentence are particularly useful in determining the overall sentiment of a document and play a crucial role in selecting the best features.

Like many other languages, colloquial Persian is prevalent in social networks, which presents challenges in processing the Persian language due to its unique features and diverse writing styles. The conversational writing style found on social networks further exacerbates the difficulties in processing textual content. Even linguistic experts may encounter challenges when dealing with everyday text from social networks and may require assistance.

Abnormal styles, such as sentence parts (verb, subject, object, etc.) being deleted or their order changed (e.g., "رفتم خونه ی دوستم" instead of "من به خانه ی دوستم رفتم," equal to "I went to my friend's house" in English), as well as the abnormal repetition of letters (e.g., "لااایک," equal to "like" in English), and the use of misspelled words (e.g., "حتا," equal to "even" in English), are often found in social network texts.

While there are powerful tools for POS tagging in English, some languages, such as Persian, still need a comprehensive tool for this purpose, particularly for colloquial Persian. In recent years, there have been attempts to develop integrated Persian preprocessing packages, with notable examples being "Hazm" [8] and "ParsiPardaz", [9] both of which are almost complete and open-source.

The flexibility of the Persian language presents another challenge in POS tagging, making it more difficult than in other languages [10]. The creation of. Creating new terms in Persian is effortless, and prefixes and suffixes can easily combine with other words to generate new terms. Persian is considered a morphologically rich language [11]. Therefore, it is possible to make up, allowing for the formation of different terms by changing affixes. For instance, the word "آمد" (which means "(s)he came") is a verb. When combined with "در," the new word is "درآمد," which means "income" with the noun tag. Similarly, if it is combined with the prefix "کار," the resulting word is "کارآمد," meaning "efficient" with an adjective tag [12]. On the other hand. However, this morphological richness also adds complexity to POS tagging, as the correct tag needs to be assigned based on the word's context.

Furthermore, distinguishing Persian text content from languages like Arabic, Urdu, and Pashto is challenging, and processing extracted colloquial text content presents difficulties. Hazm defines POS tags for formal tokens well but may need to be more effective for informal texts. Additionally, words and phrases in language can evolve, especially with the rapid growth of social networks. New terms and expressions constantly enter the Persian language due to the expansion of the internet such as "می‌چتیم" (in English: we are speaking).Hence, it is imperative to note that conventional tools like Hazm may be unreliable for part-of-speech (POS) tagging within social network analysis. This limitation stems from the fact that the underlying model of such tools is primarily trained on formal textual sources, such as official news articles and newspapers. Furthermore, the Persian popular corpus, exemplified by BijanKhan[13], which has historically served as a valuable resource for numerous studies in Persian text processing, predominantly comprises data derived from daily news and common textual materials. However, it is worth emphasizing that this corpus may no longer align with contemporary research demands, particularly in the context of POS tagging for informal vernacular expressions encountered on platforms like Twitter or Instagram.

Automatic POS tagging is a crucial step in NLP pipelines, but it requires a significant amount of annotated data to train reliable models. Social media texts differ greatly from formal texts in terms of grammar, spelling variations, slang, and abbreviated expressions, especially in Persian. Existing Persian corpora are valuable for analyzing formal text and POS tagging. Still, they may need to be more suitable for processing social media texts for applications like intelligent advertising or recommender systems.

The lack of a sufficiently large reference corpus of social media texts for training POS taggers could be why automatic POS tagging for social media texts has rarely been studied in the Persian language. Building labeled corpora, including part-of-speech tagging, presents numerous challenges, such as gathering data from diverse sources like Telegram, Twitter, and Instagram, each with its methods and techniques. Accessing large volumes of data for research purposes can be difficult, and setting up data collection tools to extract such data can be time-consuming and expensive.

Despite the significant research in this area, more work must be done on colloquial POS tagging in Persian. In this study, we aim to address this gap by introducing a comprehensive corpus called

CPPOS (Colloquial Persian Part of Speech), designed to tackle the challenges associated with informal text. CPPOS contains formal and informal sentences from three social media platforms: Telegram, Twitter, and Instagram. The data collection process spans from June 22, 2019, to March 20, 2021, to ensure coverage of various topics.

After data collection, a preprocessing step is conducted to clean the text by removing links, extra symbols, etc. The cleaned text is then tokenized using a Persian tokenizer, making it ready for annotation. Before annotation, three Persian linguistic experts created an annotation guideline, including general and specific rules and examples for specific cases. This guideline is continuously updated and is a fundamental reference for current annotators and possible future expansions of labeled data. The annotation process spans six months and involves 520K tokens, and the tagset comprises more than 60 tags.

Finally, based on this annotated daظـtaset, we train a BiLSTM model for automated POS labeling. Our contributions to this research can be summarized as follows:

- Providing an annotation guideline for colloquial Persian POS tagging.
- Introducing a novel colloquial Persian corpus called CPPOS, collected from social media platforms (Telegram, Twitter, and Instagram), consisting of over 520k tokens of both formal and informal text, including everyday phrases, encompassing more than 60 tags.
- Development of an automated labeling model based on this corpus.

The rest of the study is organized as follows: The next section reviews previous works on POS tagging. The details of the CPPOS corpus are then presented in the section titled "Corpus Preparation." The subsequent section describes the proposed method. The section on Evaluations and Results contains the performance analysis and experimental results. Finally, the conclusions and future work are presented.

## 2. Related Work

A literature review reveals that studies can be classified from different perspectives. Some studies focus solely on preparing a corpus for POS tagging, while others aim to develop automated POS labeling tools.

One of the most important corpora in English is the Penn Treebank [14], which consists of over 4.5 million words of American English. This corpus has been semi-automatically annotated for POS information over more than 3 years. Peykare (also known as the Bijankhan corpus) [15] is the most famous corpus in Persian, collected from various sources such as daily news dissertations, books, magazines, and weblogs. Written and spoken texts were systematically gathered from a diverse pool of 68 distinct subjects. This meticulous selection encompassed various lexical and grammatical structures, ensuring comprehensive coverage within the collected data.The methodology used in the annotation of Peykare involved a semi-automatic approach. Peykare contains approximately 10 million tagged words with a tag set that comprises 109 Persian POS tags. Additionally, a subset of this collection has been tagged with a smaller tag set.

Numerous POS tagging methods have already been presented, and each one has been applied to achieve high performance and accuracy on formal text. SVMTool [16] is one of the basic tools for the English language, and it has achieved an accuracy of about 98% for POS tagging in English and German. However, the tagging problem still requires further attention [17], and several unanswered questions remain. Manning [18] highlighted some of the remaining challenges in POS tagging, considering parameters such as training data size, tagging mechanism, language type, and implementation method, which can all affect a tagger's performance. Halácsy et al. [10] [19] proposed a reimplementation of the TnT POS tagger [20], which works more precisely and quickly due to better and more efficient implementation.

POS tagging solutions can be categorized into two distinct approaches: firstly, rule-based algorithms, exemplified by [18] [17]; and secondly, intelligent approaches, including Artificial Support Vector Machine (SVM) [19], Neural Networks (ANN) [20] [21] [22] [23], and Evolutionary algorithms [10].

The first serious attempt to develop a POS tagging system in Persian was made by Assi et al. [21]. The system was based on the hypothesis that syntactic behavior is reflected in co-occurrence patterns and was designed as part of the annotation procedure for a Persian corpus called The Farsi Linguistic Database (FLDB). The similarity between two words is measured concerning their syntactic behavior to their left side by the degree to which they share the same neighbors on the left. The method used in this project did not allow for a large tagset because relations among nonadjacent words were not considered in the distributional method. Consequently, they used a tagset comprising 45 tags, and the accuracy of the automatic part of the system proved to be 57.5%.

Miangah presented a method for POS disambiguation of Persian texts [24], using word class probabilities extracted from a relatively small training corpus to tag unrestricted Persian texts automatically. The process of POS tagging required passing through two levels of unigram and bigram genotype disambiguation. The results from the two levels showed that using the immediate right context of the given word significantly increased the system's accuracy rate.

Later, Miangah et al. [25]introduced a model known as the Iterative Improved Feedback model used to disambiguate Persian words in terms of POS. They developed a heuristic model that receives a raw corpus of Persian and all possible tags for every word, referred to as genotypes. Subsequently, they applied a mathematical heuristic model to extract the relevant frequencies of the word genotypes from the raw texts. During the tagging process, the algorithm passed through several iterations corresponding to n-gram levels of analysis to disambiguate each word based on a previously defined threshold. The total accuracy was calculated as 93%, which is encouraging for tagging Persian texts. However, this method required a large annotated Persian corpus.TagPer was developed by Seraji [26] for Persian using the statistical part-of-speech tagger HunPoS [19], an open-source reimplementation of TnT [20] based on the Hidden Markov Model (HMM) with trigram language models. Experiments yielded an overall accuracy of 96.9% for Persian when applied to the Bijankhan corpus. Similarly, Rezai et al. [27] evaluated the TnT tagger on Persian text. The experimental results on Persian text showed that TnT provided an overall tagging accuracy of 96.64%, with 97.01% accuracy on known words and 77.77% on unknown words.

Kardan et al. [28] proposed a POS tagger tool using a two-step method. In the first step, context and tag features of the sliding window in sentences, affix, and orthographic features of each word, along with other heuristic features, were extracted from the Bijankhan corpus. In the second step, the machine learning classifier, i.e., maximum entropy, was applied to tag parts of speech. An evaluation of the proposed method on BijanKhan illustrated that the proposed method outperformed other approaches in the same Persian corpus.

Pakzad et al. [29] attempted to utilize a joint model for Persian and English languages using Corbit software. They optimized the model's features and concurrently improved its accuracy. Corbit software applies a transition-based approach for word segmentation, POS tagging, and dependency parsing. They conducted experiments on the Persian Syntactic Dependency Treebank and Universal Dependencies English Web Treebank v1.0 for Persian and English Treebanks, respectively. In this research, the joint accuracy of POS tagging and dependency parsing over the test data on Persian reached 85.59% for coarse-grained POS and 84.24% for fine-grained POS. In another research, Badpeima et al. [30] presented a statistical-based method for Persian POS tagging. In this method, normalization was performed as a preprocessing step, and then the frequency of each word was estimated as a fuzzy function for the corresponding tag. Subsequently, the fuzzy network model was formed, and the weights of each edge were determined utilizing a neural network and a membership function. This research aimed to address the challenge of probabilistic methods when dealing with limited data, where the estimates made by these models need to be corrected.

Besharati et al. [31] applied artificial neural networks, such as multi-layer perceptron (MLP) and long short-term memory (LSTM), to POS tagging to overcome the challenge of predicting the correct tags for both in-vocabulary and out-of-vocabulary words. They proposed a hybrid model, which combined the HMM and a single-layer bidirectional LSTM model, as an innovative approach to POS tagging. This hybrid model significantly improved the performance of HMM and neural models, resulting in an accuracy of 97.29% on the Bijankhan corpus. Mohtaj et al. [32] introduced a preprocessing toolkit called Parsivar, which included POS tagging and various other activities. They incorporated Maximum Entropy (ME) and Conditional Random Fields (CRF), known for achieving successful results in sequence labeling problems such as POS tagging [33, 34]. Like the previous researchers, they also utilized the Bijankhan corpus. As a result, they reported an accuracy of 95% at both word and sentence levels. They combined the probabilistic and statistical features and information obtained with a long short-term memory (LSTM) neural network, demonstrating longer-term memory. Koochari et al. [35] further trained and tested their model, achieving an accuracy rate of 98.1% based on the Bijankhan corpus.

The literature review indicates that researchers in the Persian language have utilized the BijanKhan corpus under optimal conditions for POS model development and training. However, applying such POS taggers to colloquial texts results in a significant performance loss because they have been trained on a formal corpus.

## 3. Corpus Preparation

In this section, we will explain the steps related to the preparation of CPPOS. Figure 1 illustrates the corpus preparation steps and the model development process. After data collection, we preprocess the data through a normalization step. Then, tokenization is applied to convert the text into consecutive tokens. In the next step, tokens are annotated by three Persian linguistic experts based on a predetermined labeling guideline. Additionally, a review process was conducted on the tags concurrently. Subsequently, the corpus was utilized to train deep learning models to evaluate the effectiveness of the labels.

.

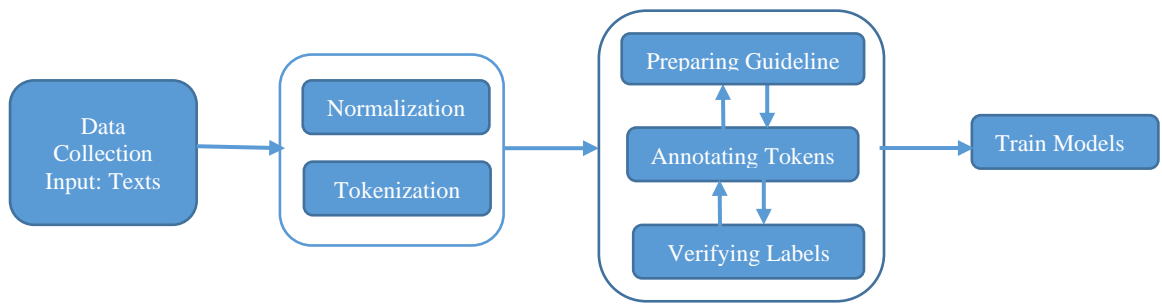

Figure 1: Steps for preparing CPPOS corpus

### 3.1. Data Collection

Data collection is the initial step that gathers and provides data for labeling. The CPPOS corpus includes texts from Telegram, Twitter, and Instagram, including posts and comments. To accomplish this, posts are collected from the target social networks using offered APIs or scraping methods if the target social network is accessible via the web (e.g., Instagram). We utilized Twitter's advanced search API to collect Persian text content from the platform. Similarly, posts published in Telegram channels were gathered using the Telegram API. The data collection period spans from June 22, 2019, to March 20, 2021. This time frame was chosen for two primary reasons:

1. Most previous Persian corpora are related to periods before 2010, rendering the related models inadequate for current requirements, given the diverse topics prevalent in social networks.

2. The emergence of COVID-19 in Wuhan, China, in December 2019 significantly impacted language, leading to the creation of new words, phrases, and literature in the news and daily conversations among people. The COVID-19 pandemic gave rise to new sentences and colloquial phrases that did not exist before. For example, "کرونایی" (referring to a person who contracted COVID-19) is a novel adjective that emerged due to the COVID-19 epidemic. Therefore, focusing on the abovementioned period was also intended to encompass topics related to COVID-19.

*3.2. Text Preprocessing*

Despite the efforts to gather Persian text content from Telegram, Twitter, and Instagram, a refinement step was necessary to identify Persian texts and potential non-Persian text, such as Arabic, Urdu, or Pashtu, which the APIs may have provided incorrectly.

Furthermore, social networks often contain duplicate or near-duplicate topics published by users. We employ Minhash-LSH[36]to address this to remove duplicate or near-duplicate texts. Manually POS labeling is expensive, so we proceed with text normalization after the duplicate detection. The normalization process includes the removal of links and emojis. However, given the significance of punctuation in the labeling process, punctuation marks are retained and not deleted during normalization. Subsequently, we tokenize the text into sentences using various delimiters. Sample delimiters are shown in Table 1. Each sentence is tokenized separately using spaces or comma characters to form a sequence of tokens.

Table 1: Sample delimiters

| Sentence Delimiters | Word Delimiters |
|---|---|
| . | Whitespace |
| ? | ( |
| ؟ | ) |
| ! | « |
| :\n | » |

Table 2 shows the number of tokens resulting at the end of the preprocessing step.

Table 2: Size of CPPOS

| **Domain** | **Number of tokens** |
|---|---|
| Twitter | 170K |
| Telegram | 200K |
| Instagram | 150K |
| CPPOS | 520K |

*3.3. POS Tagging Guideline*

The CPPOS tagset consists of more than 60 tags, several of which are novel in Persian. As previously mentioned, former researchers have primarily focused on colloquial text for POS tagging. Table 3 displays the significant tags present in the CPPOS corpus. This paper shares the lessons and experiences acquired during the collaborative labeling of 520K POS tags with linguistic experts. To ensure precise preparation of the corpora, several protocols and rules were defined to integrate the opinions of linguistic experts. Over time, these rules have been modified and refined. Some of the rules in the regulation are as follows:

1. Phrasal Verbs: Tagging phrasal verbs in the Persian language poses a challenge. Therefore, we added these verbs to the regulation over time to achieve coordination among linguists. For example, "تحویل دادن" (meaning "deliver") is one of the phrasal verbs. Detecting these verbs is difficult, even for linguists.

2. Kasre Ezafe Detection: Kasre Ezafe can be crucial in sentence parsing and word tagging in the Persian language [37]. To address this, we use a unique tag for words that end with Kasre Ezafe, adding "E" to the end of these tags. For instance, in the phrase "گل زیبا," the term "گل" ends with Kasre Ezafe, and thus, it is tagged as "NE."

3. Composite Verbs: Some verbs, like "می روم," are composed of two parts. We tag the first part with "VS," where the "S" indicates the continuation of the tag in the second part of the verb. However, the second part of these verbs does not have "S." This procedure can also be applied to nouns and adjectives. For example, some nouns (and adjectives), like "رئیس جمهور," function as a single noun. In such cases, we use the same rule: the first part gives an "S" at the end of the tag, while the second does not.

4. Unknown Tag: It is acceptable that some words may not be assigned any specific tags, especially in social media conversations. We designate these tags as the "UNK" tag in such cases.

Table 3: Some important tags in CPPOS corpus

| Tags | Description |
| --- | --- |
| N | Noun |
| V | Verb |
| ADJ | Adjective |
| ADV | Adverb |
| PRO | Pronoun |
| INTER | Interjection |
| DET | Determiner |
| PREP | Preposition |
| POSTP | Postposition |
| PUNC | Punctuation |
| HASH | Token that has been started by the hashtag sign (#) |
| CONJ | Conjunction |
| NUM | Number |
| NS | Noun that will continue in the next token |
| VS | Verb that will continue in the next token |
| ADJS | Adjective that will continue in the next token |
| ADV | Adverb that will continue in the next token |
| PROS | Pronoun that will continue in the next token |
| DETS | DET that will continue in the next token |
| PREPS | PREP that will continue in the next token |
| POSTPS | POSTP that will continue in the next token |
| CONJS | Conjunction that will continue in the next token |
| INTERS | Interjection that will continue in the next token |
| NE | Noun that ends with the Kasre |
| VE | Verb that ends with the Kasre |
| ADJE | Adjective that ends with the Kasre |
| ADVE | Adverb that ends with the Kasre |
| PROE | Pronoun that ends with the Kasre |
| INTERE | Interjection that ends with the Kasre |
| DETE | Determiner that ends with the Kasre |
| PREPE | Preposition that ends with the Kasre |
| POSTPE | Postposition that ends b with the Kasre |
| CONJE | Conjunction that ends with the Kasre |
| NSE | Noun that will continue in next token and ends by the Kasre |
| N+N | Writer has stuck two nouns without the space or with the half-space |
| NE+NS | Writer has sticked the noun with Kasre and the noun that will continue in the next token without the space or with the half-space |
| PREP + ADJ | Writer has sticked the preposition and adjective without the space or with the half-space |
| N+PREP | Writer has sticked nouns and preposition without space or with half-space |
| N+ADJ | Writer has sticked noun and adjective without space or with half-space |
| N+NS | Writer sticks two nouns |
| N+PRO | Writer has sticked noun and pronoun without space or with half-space |
| PREP+ADJ | Writer has sticked preposition and adjective without space or with half-space |
| CONJ+ADJ | Writer has sticked conjunction and adjective without space or with half-space |
| DET+ADJ | Writer has sticked determiner and adjective without space or with half-space |
| ADV+V | Writer has sticked the adverb and verb without the space or with a half-space |
| ADJ+V | The writer has sticked the adjective and verb without the space or with a half-space |
| PRO+N | The writer has sticked the pronoun and noun without the space or with a half-space |

Conversations in social media possess unique properties that make developing efficient POS tagging very challenging. Some of these properties include:

Novel Words: Some words did not exist in the Persian language before the advent of social media messengers, such as "زنگیدم" (meaning "I called you").

Sticking Words: Iranian users sometimes write words without spaces (whether consciously or unconsciously), such as "بایدبرویم" (we must go). As shown in Table 3, we consider the "+" in the POS tag when users stick at least two words without space or with a half-space.

Colloquial Words: Colloquial language is a linguistic style used for informal communication. It is commonly observed in informal texts, especially conversations between social network users.

Failure to Follow Grammar Rules:

Users of social media messengers often disregard established grammatical conventions when writing. This poses a significant challenge for part-of-speech (POS) tagging, representing a particularly troublesome scenario. Consequently, the absence of prior research addressing this issue in the context of the Persian language becomes evident. Given the compilation of the CPPOS corpus, which encompasses informal textual content, we assert our capability to address and resolve this problem. The ensuing results substantiate our claim.

*3.4. Statistics from CPPOS*

Table 4 shows the frequency of tags in the CPPOS corpus. As the table shows, the noun is the most common tag, accounting for 30% of the corpus tokens. Tags N and NE (representing nouns and the Ezafe construction [37]) indicate that the related token is a type of noun.

Table 4: Frequency of tags in CPPOS

| Domain | Number of tokens |
|---|---|
| N | 18 |
| NE | 13.4 |
| V | 12.9 |
| PUNC | 10.5 |
| PREP | 7.9 |
| ADJ | 5.7 |
| ADV | 5 |
| VS | 4.9 |
| NUM | 3.3 |
| CONJ | 3 |
| DET | 3 |
| PRO | 2.6 |
| O | 9.8 |

## 4. Experimental Results

This section discusses the various models applied to the CPPOS corpus. Section 4.1 covers the word embedding methods, while in section 4.2, we present the different models used for POS tagging.

*4.1. Word Embedding*

Representing words and documents is one of the most essential NLP tasks [38]. Word embedding involves a learned representation of text, where words with similar meanings possess comparable representations. Several approaches exist for extracting embeddings from words, and in this study, we utilize three popular and efficient word embedding methods. These three embeddings are compared in the results section and comprise ParsBERT [39], Multilingual Bert [40], and Glove [41]. The Glove model has been trained on the CPPOS corpus, while the other two approaches are pre-trained. We utilize the pre-trained models available in Huggingface.

*4.2. Applied Models*

This section discusses using different models, including RNN and BiLSTM, to train the POS tagging models. Given a sentence $w_1, w_2, w_3, ..., w_n$ with tags $y_1, y_2, ..., y_n$, BiLSTM predicts the probability distribution of tags for each word. The process is illustrated in Figure 1, where w_i represents the word representation vector.

For calculating the loss of an example, we employed the categorical cross-entropy loss function, which computes the following sum:

$$\text{Loss} = - \sum_{i=1}^{output\ size} y_i \log \hat{y}_i \quad (1)$$

The variable ŷ_i represents the i-th scalar value in the model output, and it corresponds to the target value. The output size refers to the number of scalar values in the model output. This loss function serves as a reliable measure of the distinguishability between two discrete probability distributions. As a result of choosing the cross-entropy loss function, we utilize Softmax as the activation function.

In our approach, we don't solely rely on the LSTM model to learn tags. Instead, we incorporate Hazm, one of the popular Persian POS tagging tools. Additionally, we apply an RNN model to the CPPOS corpus for further analysis and comparison.

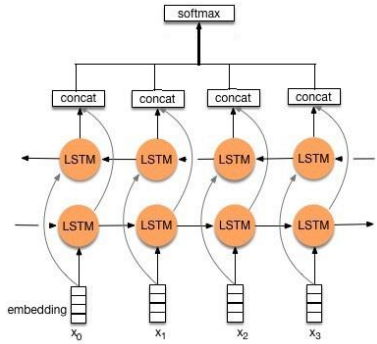

Figure 2: Bidirectional long short-term memory (BiLSTM) neural network structure [42]

Table 5 shows the various models that have been used in this paper. All the mentioned approaches shown in Table 5 are tested on the three-word embedding methods.

Table 5: Different models used in this study

| Method | Description |
| --- | --- |
| Hazm | One of the most popular Persian POS tagging tools. |
| LSTM | LSTM model |
| BiLSTM | BiLSTM model |
| RNN | RNN model |

*4.3. Environments*

We developed and tested the work in the following conditions:

- Programming language: Python3.6

- Pytorch

- Running code on the GPU (Geforce GTX 1080)

*4.4. Results*

After gathering data,and preprocessing the data, we proceeded to evaluate various models. Specifically, we employed three neural models: RNN, LSTM, and BiLSTM, all trained on the CPPOS corpus. Additionally, we utilized different word embeddings, including ParsBERT, MBERT, and Glove, and compared the performance of these models with the Hazm tool.

Table 7 displays the accuracy of the models based on the three-word embedding approaches. The BiLSTM model achieved the highest accuracy of 90.42% when using ParsBERT as the word embedding. Conversely, as indicated in Table 7, Glove yielded lower accuracy across all embedding models. Comparing Hazm and BiLSTM (the best model), Table 8 illustrates how significantly the choice of the corpus can impact the results. While Hazm, trained on the formal Bijankhan corpus, performed poorly in informal conversations, the BiLSTM model excelled.

Figure 5 reveals that the size of the embeddings directly affects network accuracy. An increase in dimensionality corresponds to an increase in accuracy up to an optimal size of 200. However, beyond 200, further increases in dimensionality can lead to a decline in model performance. As shown in Table 6, the best hyperparameters were determined through experimentation and subsequently fixed for the prediction phase. Additionally, we employed early stopping, and terminating training at epoch 20, which produced the best model.

Figure 6 demonstrates two example sentences fed into our model and Hazm. Tables 9 and 10 present the prediction performance of the two models. As evident in Tables 9 and 10, the BiLSTM model on the CPPOS corpus outperformed Hazm, with fewer errors and superior performance in colloquial text.

These practical examples exemplify the efficacy of our developed tool. Hazm and our developed tool did not assign the underlined tokens in the examples true POS tags. The real test cases demonstrate that our model exhibits fewer errors and performs better in colloquial text.

Table 6: Hyperparameters for training BiLSTM

| Parameters | Value |
|---|---|
| BERT pre-trained weights | HooshvareLab/bert-base-parsbert-uncased |
| Sentence length | 128 |
| Number of layers | 2 |
| Dropout | 0.5 |
| Batch size | 16 |
| Optimizer | Adam |
| Learning rate | 3e-5 |

Table 7: Evaluation of different methods by the accuracy

| | ParsBERT | | | Multilingual Bert | Glove |
|---|---|---|---|---|---|
| | 50D | 100D | 200D | | 100D |
| RNN | 75.36 | 81.12 | 80.95 | 82.03 | 80.75 |
| LSTM | 77.33 | 73.47 | 88.02 | 84.95 | 83.49 |
| BiLSTM | 80.95 | 85.98 | **90.42** | 87.14 | 85.69 |

Table 8: Hazm vs BilSTM

| | |
|---|---|
| Hazm (Bijankhan corpus) | 76.58 |
| BiLSTM(CPPOS corpus) | 90.42 |

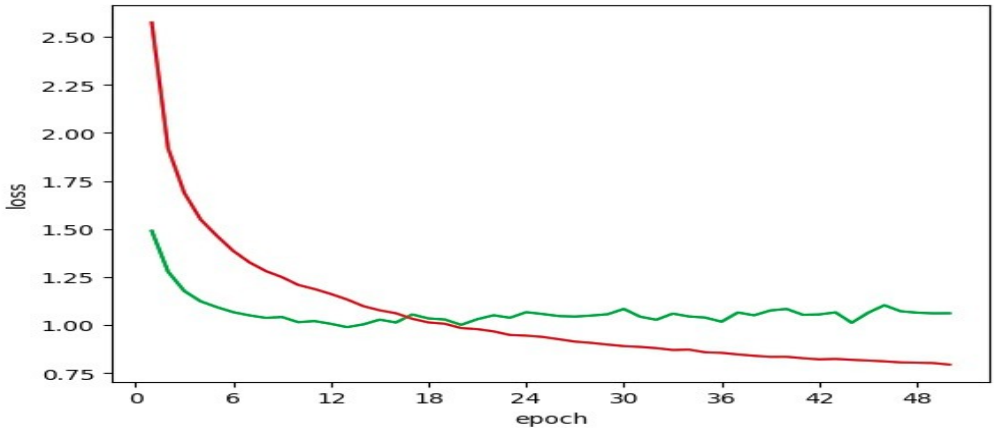

Figure 6: Early stopping at epoch 20, the green line shows evaluation loss, and the red line shows training error.

Table 9: A Real test case

| Input | خب | منم | امروز | میرما |
|---|---|---|---|---|
| Hazm Model | ADV | V | ADV | N |
| BiLSTM Model (on CPPOS) | INTER | PRO | N | V |

Table 10: A Real test case

| Input | فک | کردی | ماسک | بزنی | کرونایی | نمیشی |
|---|---|---|---|---|---|---|
| Hazm Model | N | V | N | V | NE | AJ |
| BiLSTM Model (on CPPOS) | VS | V | N | V | N | V |

## 5. Conclusion

The POS tagger is crucial for many NLP tasks, such as text summarization and machine translation. However, in Persian, no corpus includes social media conversations, and the available ones and tools for POS tagging do not support tagging slang or colloquial phrases. Developing a model that can tag formal and informal text simultaneously poses a significant challenge, and research in this area is yet to be explored. POS tagging for texts on popular Persian applications like Telegram, Twitter, and Instagram is essential for natural language processing in the Persian language.

In this study, we constructed the CPPOS corpus, including formal and informal texts. The CPPOS corpus, consisting of over 520K samples, was gathered from Telegram, Twitter, and Instagram applications. It is the first Persian corpus that simultaneously incorporates formal and informal text. After collecting the corpus, we developed various models for POS tagging classification, such as RNN, LSTM, and BiLSTM. The best results were achieved with two layers of BiLSTM, followed by the Softmax layer after the BiLSTM layers. Word embeddings used in this project included ParsBERT, MBERT, and Glove, with the best results obtained using ParsBERT.

The main advantage of this paper lies in introducing the CPPOS corpus and utilizing the BiLSTM model to uncover hidden information. The results demonstrated that the BiLSTM model successfully predicted POS tagging for both formal and informal texts simultaneously. Furthermore, we showed that our model, trained on the CPPOS corpus, outperformed the Hazm tool.

**Conflicts of interests**

The authors did not receive support from any organization for the submitted work.

**Data availability**

The datasets generated during the current study are available from the corresponding author on reasonable request.

*References*

1. Voutilainen, A., *Part-of-speech tagging.* The Oxford handbook of computational linguistics, 2003: p. 219-232.
2. Szkoła, J., K. Pancerz, and J. Warchoł, *Recurrent neural networks in computer-based clinical decision support for laryngopathies: an experimental study.* Computational Intelligence and Neuroscience, 2011. **2011**: p. 7-7.
3. Chotirat, S. and P. Meesad, *Part-of-Speech tagging enhancement to natural language processing for Thai wh-question classification with deep learning.* Heliyon, 2021. **7**(10): p. e08216.
4. Aivazoglou, M., et al., *A fine-grained social network recommender system.* Social Network Analysis and Mining, 2020. **10**: p. 1-18.

5. Rajabi, Z. and M. Valavi, *A survey on sentiment analysis in Persian: A comprehensive system perspective covering challenges and advances in resources and methods.* Cognitive Computation, 2021. **13**(4): p. 882-902.
6. Rajabi, Z., M.R. Valavi, and M. Hourali, *A context-based disambiguation model for sentiment concepts using a bag-of-concepts approach.* Cognitive Computation, 2020. **12**: p. 1299-1312.
7. Shamsfard, M., *Challenges and Opportunities in Processing Low Resource Languages: A study on Persian.*
8. *Python library for digesting Persian text.* 2014; Available from: Https://github.com/sobhe/hazm.
9. Sarabi, Z., H. Mahyar, and M. Farhoodi. *ParsiPardaz: Persian language processing toolkit.* in *ICCKE 2013*. 2013. IEEE.
10. Hosseini Pozveh, Z., A. Monadjemi, and A. Ahmadi, *Persian texts part of speech tagging using artificial neural networks.* Journal of Computing and Security, 2016. **3**(4): p. 233-241.
11. Perry, J.R., *Persian morphology.* Morphologies of asia and africa, 2007. **2**: p. 975-1019.
12. Passban, P., Q. Liu, and A. Way, *Boosting neural POS tagger for Farsi using morphological information.* ACM Transactions on Asian and Low-Resource Language Information Processing (TALLIP), 2016. **16**(1): p. 1-15.
13. BijanKhan, M., *The Role of the Corpus in Writing a Grammar: An Introduction to a Software.* Iranian Journal of Linguistics, 2004. **19**.
14. Marcus, M., B. Santorini, and M.A. Marcinkiewicz, *Building a large annotated corpus of English: The Penn Treebank.* 1993.
15. Bijankhan, M., et al., *Lessons from building a Persian written corpus: Peykare.* Language resources and evaluation, 2011. **45**(2): p. 143-164.
16. Giménez, J. and L. Marquez. *SVMTool: A general POS tagger generator based on Support Vector Machines.* in *In Proceedings of the 4th International Conference on Language Resources and Evaluation.* 2004. Citeseer.
17. Giesbrecht, E. and S. Evert. *Is part-of-speech tagging a solved task? An evaluation of POS taggers for the German web as corpus.* in *Proceedings of the fifth Web as Corpus workshop.* 2009. Citeseer.
18. Manning, C.D. *Part-of-speech tagging from 97% to 100%: is it time for some linguistics?* in *Computational Linguistics and Intelligent Text Processing: 12th International Conference, CICLing 2011, Tokyo, Japan, February 20-26, 2011. Proceedings, Part I 12.* 2011. Springer.
19. Halácsy, P., A. Kornai, and C. Oravecz, *HunPos-an open source trigram tagger.* 2007.
20. Brants, T., *TnT-a statistical part-of-speech tagger.* arXiv preprint cs/0003055, 2000.
21. Assi, S.M., *Grammatical tagging of a Persian corpus.* International journal of corpus linguistics, 2000. **5**(1): p. 69-81.
22. Mohseni, M. and B. Minaei-Bidgoli. *A Persian Part-Of-Speech Tagger Based on Morphological Analysis.* in *LREC.* 2010.
23. Bijankhan, M., *The role of the corpus in writing a grammar: An introduction to a software.* Iranian Journal of Linguistics, 2004. **19**(2): p. 48-67.
24. Tayebeh, M.M., *Corpus-based part-of-speech disambiguation of Persian.* 2011.

25. Miangah, T.M. and A.D. Khalafi, *Unsupervised part of speech tagging for Persian.* International Journal of Artificial Intelligence & Applications, 2012. **3**(2): p. 33.

26. Seraji, M. *A statistical part-of-speech tagger for Persian*. in *NODALIDA 2011, Riga, Latvia, May 11–13, 2011*. 2011.

27. Rezai, M.J. and T. Mosavi Miangah, *FarsiTag: A part-of-speech tagging system for Persian.* Digital Scholarship in the Humanities, 2017. **32**(3): p. 632-642.

28. Kardan, A.A. and M.B. Imani. *Improving Persian POS tagging using the maximum entropy model*. in *2014 Iranian Conference on Intelligent Systems (ICIS)*. 2014. IEEE.

29. Pakzad, A. and B. Minaei Bidgoli, *An improved joint model: POS tagging and dependency parsing.* Journal of AI and Data Mining, 2016. **4**(1): p. 1-8.

30. Badpeima, M., F. HOURALI, and M. HOURALI, *Part of speech tagging of Persian Language using fuzzy network model.* 2019.

31. Besharati, S., et al., *A hybrid statistical and deep learning based technique for Persian part of speech tagging.* Iran Journal of Computer Science, 2021. **4**: p. 35-43.

32. Mohtaj, S., et al. *Parsivar: A language processing toolkit for Persian*. in *Proceedings of the eleventh international conference on language resources and evaluation (lrec 2018)*. 2018.

33. Rajani Shree, M. and B. Shambhavi, *POS tagger model for Kannada text with CRF++ and deep learning approaches.* Journal of Discrete Mathematical Sciences and Cryptography, 2020. **23**(2): p. 485-493.

34. Xuan Bach, N., T. Khuong Duy, and T. Minh Phuong. *A POS tagging model for Vietnamese social media text using BiLSTM-CRF with rich features*. in *PRICAI 2019: Trends in Artificial Intelligence: 16th Pacific Rim International Conference on Artificial Intelligence, Cuvu, Yanuca Island, Fiji, August 26-30, 2019, Proceedings, Part III 16*. 2019. Springer.

35. Koochari, A., A. Gharahbagh, and V. Hajihashemi. *A Persian part of speech tagging system using the long short-term memory neural network*. in *6th Iran. Conf. Signal Process. Intell. Syst. ICSPIS*. 2020.

36. Manaa, M.E. and G. Abdulameer, *Web Documents Similarity using K-Shingle tokens and MinHash technique.* J. Eng. Appl. Sci, 2018. **13**: p. 1499-1505.

37. Asghari, H., J. Maleki, and H. Faili. *A probabilistic approach to persian ezafe recognition*. in *Proceedings of the 14th Conference of the European Chapter of the Association for Computational Linguistics, volume 2: Short Papers.* 2014.

38. Wang, S., W. Zhou, and C. Jiang, *A survey of word embeddings based on deep learning.* Computing, 2020. **102**: p. 717-740.

39. Farahani, M., et al., *Parsbert: Transformer-based model for persian language understanding.* Neural Processing Letters, 2021. **53**: p. 3831-3847.
40. Libovický, J., R. Rosa, and A. Fraser, *How language-neutral is multilingual BERT?* arXiv preprint arXiv:1911.03310, 2019.
41. Pennington, J., R. Socher, and C.D. Manning. *Glove: Global vectors for word representation.* in *Proceedings of the 2014 conference on empirical methods in natural language processing (EMNLP)*. 2014.
42. Limsopatham, N. and N.H. Collier, *Bidirectional LSTM for named entity recognition in Twitter messages.* in Proceedings of the 2nd Workshop on Noisy User-generated Text (WNUT). 2016.26